\documentclass[reprint,amsmath,amssymb, aps]{revtex4-2}
\usepackage{multirow}
\usepackage{xcolor}
\usepackage{natbib}
\usepackage{subcaption}
\usepackage{graphicx}
\usepackage{url}
\usepackage{fancyhdr}
\usepackage{adjustbox}

\begin{document}

\title{What's in an embedding?\\
Would a rose by any embedding smell as sweet?}
\author{Venkat Venkatasubramanian}%
\email{venkat@columbia.edu}
\affiliation{Complex Resilient Intelligent Systems Laboratory \\Department of Chemical Engineering \\Columbia University, New York, NY 10027}

\maketitle

\section*{Abstract}

Large Language Models (LLMs) are often criticized for lacking true "understanding" and the ability to "reason" with their knowledge, being seen merely as autocomplete systems. We believe that this assessment might be missing a nuanced insight. We suggest that LLMs do develop a kind of empirical "understanding" that is "geometry"-like, which seems adequate for a range of applications in NLP, computer vision, coding assistance, etc. However, this "geometric" understanding, built from incomplete and noisy data, makes them unreliable, difficult to generalize, and lacking in inference capabilities and explanations, similar to the challenges faced by heuristics-based expert systems decades ago. \\

To overcome these limitations, we suggest that LLMs should be integrated with an "algebraic" representation of knowledge that includes symbolic AI elements used in expert systems. This integration aims to create large knowledge models (LKMs) that not only possess "deep" knowledge grounded in first principles, but also have the ability to reason and explain, mimicking human expert capabilities. To harness the full potential of generative AI safely and effectively, a paradigm shift is needed from LLM to more comprehensive LKM. 

\section{Introduction}

'What's in a name? That we call a rose by any other name would smell as sweet.” These immortal words from Shakespeare's Romeo and Juliet resonate in the context of large language models. One might ask: What’s in an embedding? Would a rose by any embedding still smell as sweet? In other words, does an embedding capture all the properties of an entity comprehensively? \\

Related to this is the important question: Do LLMs truly "understand" the meaning of the words and sentences they generate as humans do, or are they "stochastic parrots"~\cite{bender2021parrots} merely performing autocomplete without any "understanding?" \\

In this paper, we address these questions from the perspective of knowledge representation~\cite{rich1983AI} in artificial intelligence. 

\section{A Tale of Two Representations: Algebra vs. Geometry}

Vector embedding of tokens in a large language model is a kind of knowledge representation.  A well-known example of the importance of knowledge representation in problem-solving is the Indo-Arabic vs. Roman numeral representation of numbers. Roman numerals are perhaps great for numbering Super Bowls, but they are a nightmare when it comes to doing math. A good knowledge representation should make features that are central to problem-solving obvious and easy to manipulate, which is what the Indo-Arabic decimal number system does for arithmetic. \\

Another interesting example is the Hindu monk problem, which is as follows. One morning, the monk sets out at sunrise to climb a path up the mountain to reach the temple at the top of the mountain. He arrives at the temple just before sunset. A few days later, he leaves the temple at sunrise to descend the mountain, traveling somewhat faster since it is downhill. Along the way, in both his upward and downward journey, he rests at different spots at different times briefly. Is there a spot along the path that the monk will occupy at precisely the same time of day on both trips? \\

Knowledge representation is the key to solving this problem. Thinking about the problem verbally or in terms of rates and distances, for example, is unproductive. It is nearly impossible to solve the problem this way. However, it is easily solved by mere inspection by representing the problem visually. Visualize the monk's image on the path ascending the mountain, starting at dawn. At the same time, visualize another image of the monk on the same path descending the mountain, also starting at dawn. Obviously, the two images will meet somewhere on the trail and, therefore, occupy the same spot at the same time of day. The problem that is virtually impossible to solve in the verbal representation is solved trivially in the visual representation. The key to this problem is knowing the location of the monk on the path at different times. The visual representation makes this central feature explicit and easy to manipulate. \\

Thus, both Indo-Arabic numerals and visual representation make important features of the problem obvious and easy to manipulate, thereby making reasoning much easier. This is central to many problem-solving tasks. \\

Now, let us consider the algebraic vs. the geometric representation of mathematical objects, such as a circle. In algebra, a circle of radius $r = 1$ whose center is at the (x,y) coordinates (0,0) is given by
\begin{equation}
    x^2 + y^2 = 1
    \label{eq:circle1}
\end{equation}

\begin{figure}[!ht]
    \begin{subfigure}{0.45\linewidth}
        \centering
        \includegraphics[width=\linewidth]{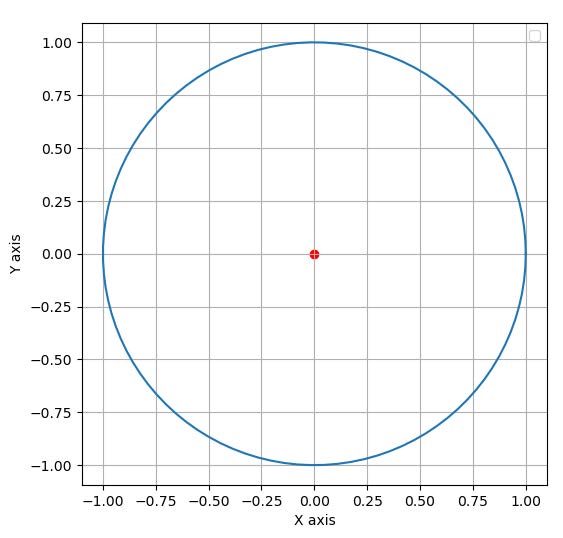}
        \caption{Circle with radius $r=1$}
        \label{fig:circle-rad1}
    \end{subfigure}
    \hfill
    \begin{subfigure}{0.45\linewidth}
        \centering
        \includegraphics[width=\linewidth]{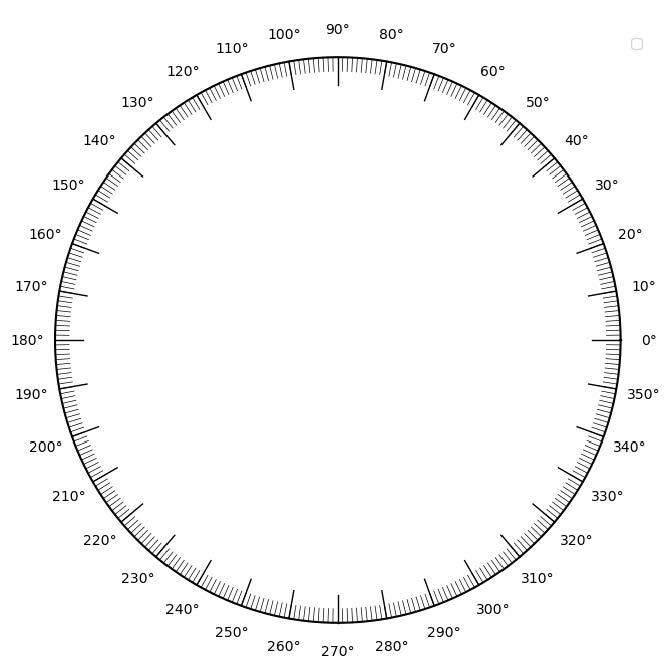}
        \caption{Circle with angular markings}
        \label{fig:circle-angles}
    \end{subfigure}
    \caption{Geometric representation of a circle}
    \label{fig:circle1}
\end{figure}

The geometric representation of the same circle is seen in Fig.~\ref{fig:circle-rad1}. In Fig.~\ref{fig:circle-angles} we see a geometric representation of a circle with the angular information ($360^{\circ}$) explicitly marked.

Now, let us perform a thought experiment. Consider two individuals: Alice, who is familiar only with the algebraic (i.e., symbolic) representation of a circle, and Georgette, who understands only the geometric representation. Now, who truly "knows" or "understands" the circle better? Whose representation is more comprehensive? \\

Well, it all depends on the questions they are asked. Consider the following questions: Are points A (0.8, 0.6) and B (-0.8, -0.6) on the circumference of the circle? Are these points close to each other or far apart? Are they on opposite ends of the circle? What is the angle subtended by the points  C (0.866, 0.5) and D (-0.342, 0.94)? \\

To answer these questions, Alice, using only algebraic reasoning, will have to first substitute these coordinates into Eq.~\ref{eq:circle1} to verify whether they lie on the circumference. Then, to find out whether they are close or far, and whether they lie at opposite ends, she will have to calculate the arc length between the points, which would require some additional computations and analytical reasoning. \\

\begin{figure}[!ht]
    \begin{subfigure}{0.45\linewidth}
        \centering
        \includegraphics[width=\linewidth]{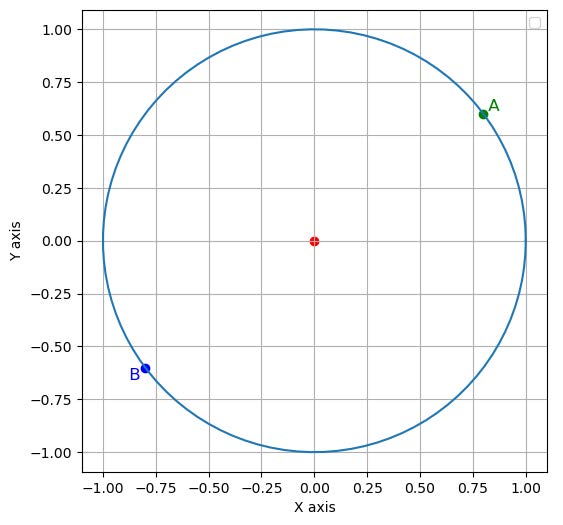}
        \caption{Points A and B on the circle}
        \label{fig:circle-AB}
    \end{subfigure}
    \hfill
    \begin{subfigure}{0.45\linewidth}
        \centering
        \includegraphics[width=\linewidth]{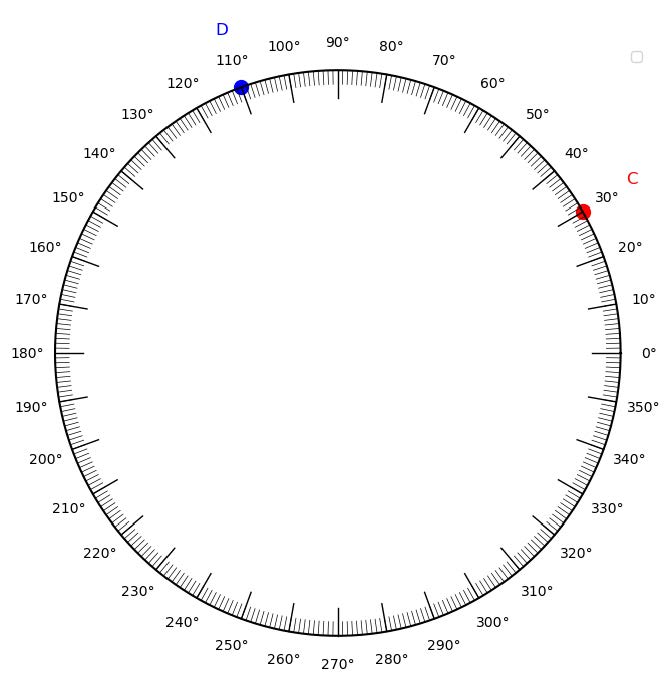}
        \caption{Points C and D on the circle}
        \label{fig:circle-CD}
    \end{subfigure}
    \caption{Points on the circle}
    \label{fig:circle2}
\end{figure}

On the other hand, Georgette would merely locate the points A and B in her geometric representation of the circle (Fig.~\ref{fig:circle-AB}), and mere inspection tells her that they are on the circumference, that they are far apart and, in fact, are on opposite ends of the circle. Using Fig.~\ref{fig:circle-CD}, she can see that the angle subtended between points C and D is $80^{\circ}$. All this requires absolutely no reasoning at all. The answers are obvious and are simply looked up. Georgette doesn't reason because she doesn't have to! The questions that require some analytical effort on the part of Alice are effortless for Georgette. Like the Hindu monk problem, the representation makes all the difference. A picture is worth a thousand words, indeed!\\

So, when one asks Georgette for an explanation about how she reached her conclusions, her answer might be something like: "Points A and B are on the circle because they are! And I can see that they are on opposite ends. There is no need to explain any of this!" She cannot explain her conclusions to anyone who does not have access to her representation. \\

Let us consider one final example. Consider two circles given by the 
\begin{eqnarray}
    x^2 + y^2 = 4 \\
    (x-1.5)^2 + (y - 1.0)^2 = 1 
    \label{eq:circles2}
\end{eqnarray}

\begin{figure}
        \centering
        \includegraphics[width=0.75\linewidth]{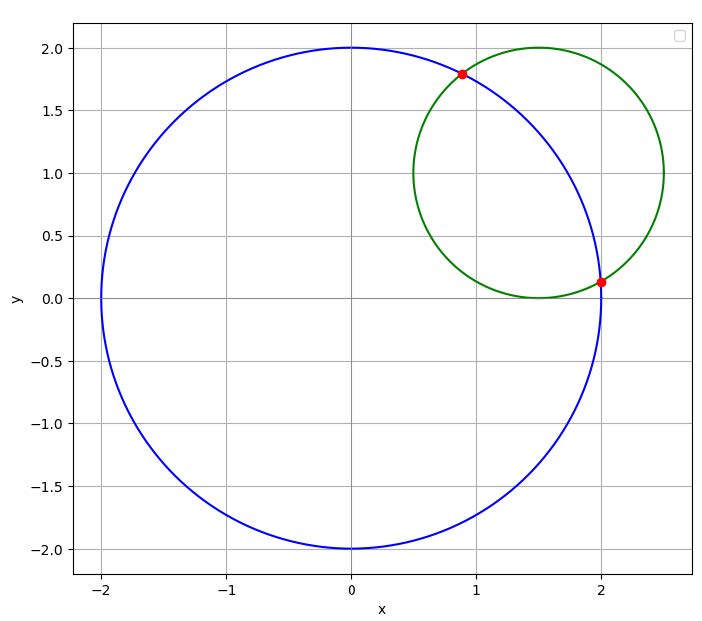}
        \caption{Intersecting circles}
        \label{fig:intersect_circles}
\end{figure}

Do these circles intersect? If so, what are the intersection points? Again, for Georgette, the answers are trivial. She sees from Fig.~\ref{fig:intersect_circles} that the intersecting points are approximately (0.89, 1.79) and (2.00, 0.13). However, to arrive at the same results using the algebraic representation, Alice will have to do some analytical reasoning and computation. \\

Thus, the "geometric rose" smells sweeter than the "algebraic rose" in these situations. \\

Now, let us turn the tables and ask a different kind of question. What is the representation of a circle in three- and higher-dimensional spaces? What are the surface area and volume of a three-dimensional circle (i.e., a sphere)? \\

As we know, these questions cannot be answered using the 2-D geometric representation of a circle in Figs.~\ref{fig:circle1}-\ref{fig:circle2}. We will need additional data to show the sphere in 3-D. Even with that, it is not easy to calculate the surface area and volume. For higher dimensions, visual representation is impossible. However, we can readily generalize the circle in Eq.~\ref{eq:circle1} to higher dimensions. Furthermore, one can use calculus to determine the surface area and volume quite easily in the algebraic representation. Thus, the algebraic representation is superior for these kinds of problems. \\

So, whose understanding is better and more comprehensive? Who has \textit{really} understood what a circle is? One could argue that, despite the expediency offered by the geometric representation, the algebraic representation captures the "essence" of a circle more comprehensively. A circle's invariant and symmetry properties are better captured in this representation, making it easier to generalize, explain and reason with, and calculate new properties. \\

However, most importantly, just because Georgette does not (and perhaps cannot) explain and justify her conclusions, it does not mean that she does not understand what a circle is. She certainly understands it in an empirical sense, but her understanding is incomplete if it is not complemented by the algebraic view. Similarly, one could argue that Alice's understanding is also incomplete if she never sees a circle geometrically. \\

Therefore, even in highly precise disciplines such as mathematics, we can have representations that are incomplete. This dichotomy also suggests the exciting possibility that there could exist other representations in high-dimensional spaces that are "richer" in capturing and revealing important mathematical relationships. This possibility reminds us of the almost magical abilities of Srinivasa Ramanujan, the Indian mathematical genius, for whom scrolls containing the most complicated mathematics would unfold in his dreams~\cite{kanigel1991ramanujan}. Ramanujan seems to have had access to an abundantly rich representation of mathematics in his mind that was more "geometric" or "visual" in nature, in the spirit of our discussion above. Therefore, many complex and startling results were obvious to him because he "saw" them to be true. 

\section{Knowledge representation in LLMs}

The enormously successful use of high-dimensional vectors for embedding tokens (e.g., GPT-3.5 uses 12,288 dimensions) raises an interesting question: Could the LLMs be using a "geometric" representation rather than an "algebraic" one for their knowledge internally? Therefore, instead of "reasoning", they merely "look up" in their internal knowledge base to answer queries, somewhat like Georgette.  \\

An LLM's performance critically depends on the quantity and quality of data used in its training. Table~\ref{table:gpt} shows the approximate sizes of the three versions of GPT. Obviously, as the LLM has more parameters and is trained on more data, its problem-solving capability grows enormously. We see that GPT-3.5 is roughly 1000x GPT-1 in parameters and "knows" roughly 100x the knowledge of GPT-1.  \\

\begin{figure*}[!ht]
    \centering
    \includegraphics[width=\linewidth]{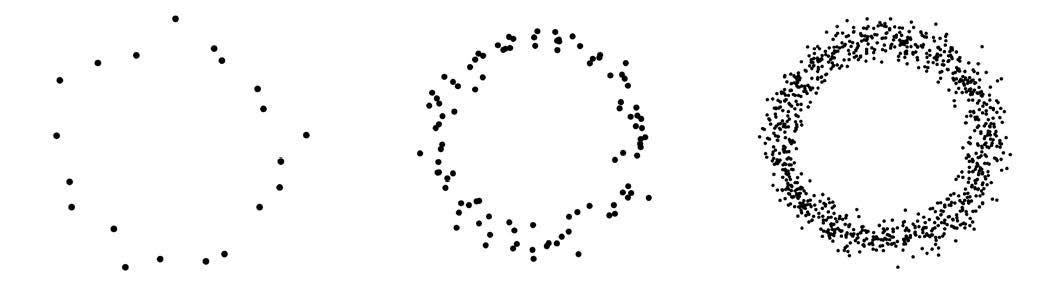}
    \caption{{ Geometric representation of a circle: (a) 20 points with noise (b) 100 points with noise (c) 1000 points with noise}}
    \label{fig:noisy-circles}
\end{figure*}

The evolution of GPT-1 to GPT-3.5 is similar to the representation of a circle with more and more noisy data points, as in Fig.~\ref{fig:noisy-circles}.  Figs.~\ref{fig:circle1}-\ref{fig:circle2} represent the perfect circle, the one with \textit{infinite noise-free} points. On the other hand, Fig.~\ref{fig:noisy-circles} represents different imperfect, incomplete, and noisy approximations of a circle. Therefore, for the aforementioned questions about points A \& B and C \& D one can get erroneous and stochastic answers depending on the approximation. Generally speaking, GPT-1 is like Fig.~\ref{fig:noisy-circles}a, whereas GPT-3.5 is like Fig.~\ref{fig:noisy-circles}c. As in Fig.~\ref{fig:noisy-circles}c, the much denser data in GPT-3.5 resulted in much better performance. \\

\begin{table}[h!]
\centering
\begin{adjustbox}{width=\linewidth,center}
\begin{tabular}{c c c c} 
 \hline
 \text{Year} & \text{Version} & \text{Parameters} & \text{Training data} \\ [0.5ex] 
 \hline
 2018 & GPT-1 & 117 M & 4.5 GB \\ 
 2019 & GPT-2 & 1.5 B & 40 GB \\
 2020 & GPT-3.5 & 175 B & 570 GB \\ [0.5ex] 
 \hline
\end{tabular}
\end{adjustbox}
\caption{Sizes of different GPT versions}
\label{table:gpt}
\end{table}

In the real world, the equivalent of the "perfect circle" is not available in most practical situations. So, what we do is develop useful approximations by training LLMs on enormous amounts of data that tend to be inherently incomplete and uncertain. This results in LLMs developing a noisy "geometric" approximation of the world. \\

Moreover, for many practical situations, the equivalent of the algebraic representation of a circle in Eq. 1 is simply not available. For example, we cannot write down such abstract equations to model natural language, vision, or game playing. Therefore, for such problems, we are simply dependent on the purely data-driven "geometric" model of the world. There are some notable exceptions in the scientific and engineering domains, which we address in the next section. \\

Recent articles from researchers in Anthropic AI~\cite{Roose2024NYT, templeton2024scaling} and Open AI~\cite{gao2024openai} shed some light on the internal representations of commercial grade LLMs. The Anthropic AI team examined one of their LLMs, Claude 3 Sonnet, using a method called "dictionary learning" to discover patterns in the activation of neuron combinations when Claude was asked to discuss specific topics. Approximately 10 million of these patterns or features were identified. \\

\begin{figure}
        \centering
        \includegraphics[width=\linewidth]{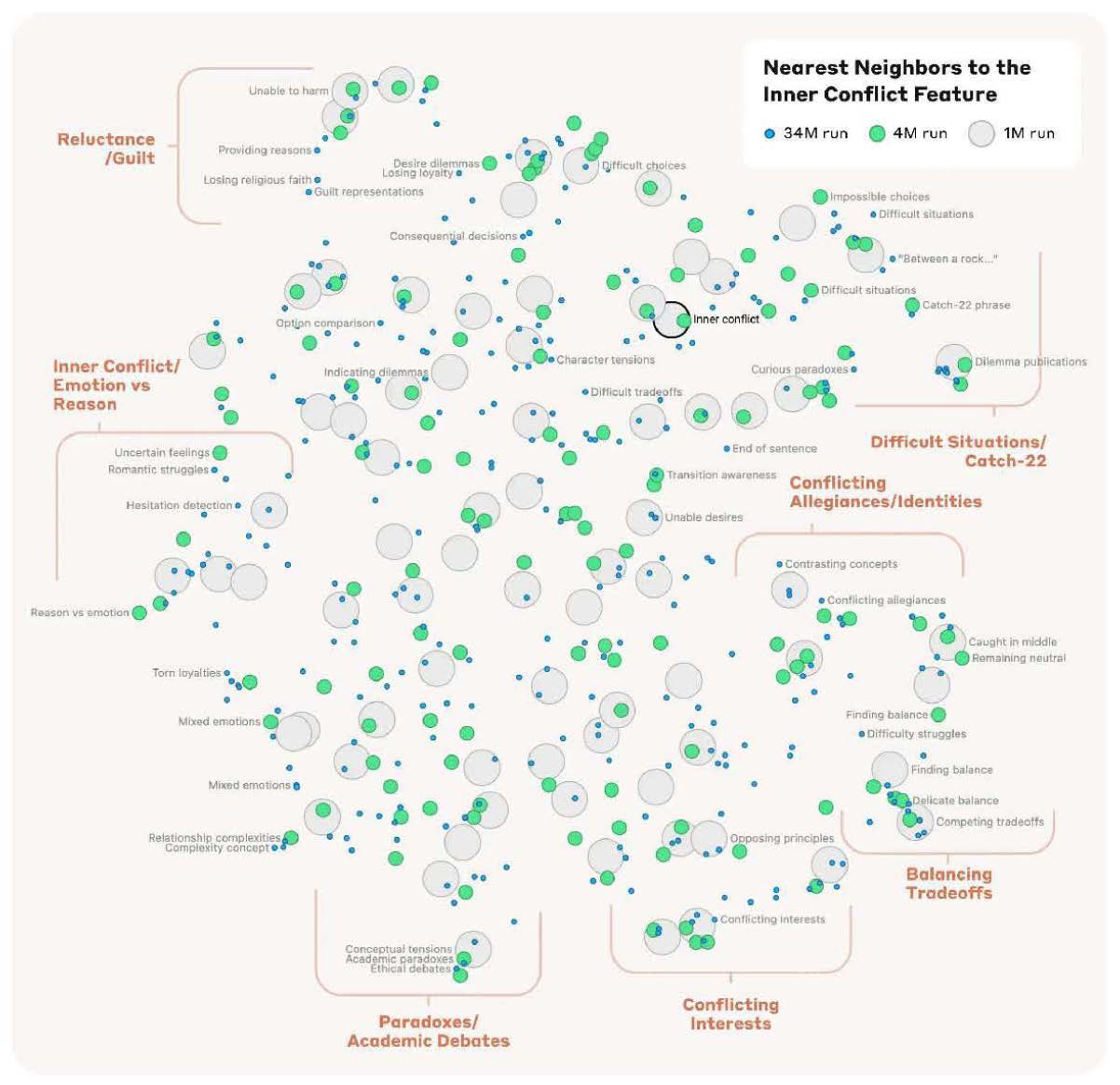}
        \caption{"Geometric" representation of features in Claude 3 Sonnet~\cite{templeton2024scaling}}
        \label{fig:claude}
\end{figure}

The team found that these features exhibit depth, breadth, and a level of abstraction indicative of Claude's sophisticated capabilities. They developed a type of "distance" measure between features based on the neurons involved in their activation patterns. This enabled them to search for features that are "close" to one another. Examining a feature related to "inner conflict", they found features associated with relationship breakups, conflicting loyalties, logical inconsistencies, and the phrase "catch-22" (Fig.~\ref{fig:claude}). This indicates that the internal organization of concepts within the AI model somewhat aligns with human notions of similarity. \\

These results resonate with our hypothesis and analysis that the LLMs are using a "geometry"-like internal representation in a high-dimensional embedding space rather than an "algebraic" one. In the circle example, the geometric representation readily revealed which points are spatially close to each other, which are further apart, what the subtended angles are, etc., to make reasoning easier. Similarly, in the high-dimensional "meanings" space represented by the vector embeddings, Claude's internal representation captured the meanings of words and phrases and their relative distances in a "geometry"-like manner, making it easier to do sophisticated "reasoning," such as analogies and metaphors. Claude "reasons" without actually doing the "reasoning" but simply "looking up." We suggest that Claude "understands" its knowledge in the sense that Georgette understands a circle. \\

An important question here is the following: What is the "algebraic" representation of Claude's knowledge? This is possible for simple geometric objects like circles, but is it even possible for complex large-scale data? If not, can we at least develop useful "analytical" approximations? \\

In this context, some of the concepts and techniques that were developed in the era of symbolic AI in the 1970s-1980s, such as rule-based representations, frames, and ontologies that were used in expert systems, could be helpful in developing such "analytical" approximations of real-world knowledge. Recent work shows how this could be done in different applications to develop mechanistic "algebraic" approximations of the world to complement the "geometric" view~\cite{venkat1988two-tier, mann2021predicting,mann2022hybrid, mann2023susie,lenat2023LLM-CYC}. \\

For someone who is born blind, it is virtually impossible to imagine what it is like to be able to see the world around him effortlessly in real time. For him, everything is located after considerable effort, through elaborate searching and feeling for objects - that is, performing spatial "reasoning." He will have no appreciation for colors, brightness, or other visual features. He will be amazed at what a person with sight is able to do effortlessly without searching - literally "look up" to locate objects and make decisions. The capabilities of a sighted person would appear magical. It is impossible to explain to him how a person with sight sees. Likewise, it is nearly impossible for us to understand how the LLMs "see" the world in a way that could be very different from ours. \\

In a biological visual system, features are extracted, assembled, and interpreted at hierarchical levels of detail. Similarly, we believe that the LLMs similarly perform "picture"-like processing and interpretations of the tokens. In this sense, the LLMs seem to possess a "mind's eye", as it were. An LLM is more like Cyclops than a stochastic parrot. Many queries that require considerable effort of searching and reasoning for us would seem obvious and easy, requiring no reasoning or search for them. Like Georgette, the LLMs do not "reason" because they don't have to! They do not "explain" because there is nothing to "explain"! \\

It is interesting to note that our theory of "geometric" and "algebraic" representations of knowledge in LLMs is similar in spirit to the cognitive theory of semantics for humans proposed by Peter G{\"a}rdenfors~\cite{gardenfors2014geometry}. 
He argues that our minds organize information in a format that can be modeled in geometric terms. We cannot "look" into the minds of people directly to test this theory; we can only resort to indirect empirical tests. However, we are fortunate to have the ability to observe the "insides" of LLMs. This is what the Anthropic AI results reveal and lend support to the "geometric" representation hypothesis. 

\section{From LLMs to LKMs for Science and Engineering}

As observed, relying only on a "geometric" understanding of the world limits the potential of LLMs, particularly for science and engineering applications. For example, a self-driving car can navigate impressively through crowded city traffic, but does it "know" and "understand" the concepts of mass, momentum, acceleration, force, and Newton's laws as humans do? \\

The car's behavior seems animal-like, such as that of a cheetah chasing an antelope in the wild. Both animals show great mastery of the dynamics of the chase, but do they “understand” the underlying physics, classical mechanics, or the underlying math, calculus? The animals do not seem to have that kind of "understanding", but they do possess a practical "understanding" of the domain, which allows them to execute the required tasks well. That is, they possess the "know-how" of different tasks, but not the "know-why." In the parlance of the expert systems era, they possess superficial "heuristic" knowledge but not "deep" knowledge~\cite{venkat1988two-tier}. \\

It is in the area of "know-why" that humans excel, at the moment, with their ability to perform abstract symbolic reasoning to develop predictive mechanistic models of the world from first principles. It appears that current LLMs have perhaps achieved animal-like mastery of their tasks, but not a "deeper" mechanistic understanding of the world, as humans do. \\

These deficiencies pose challenges for LLMs, particularly in the science and engineering domains. The scientific and engineering domains are governed by the fundamental laws of physics and chemistry (and biology), constitutive relations, and highly technical knowledge about materials, processes, and systems~\cite{venkatasubramanian2019promise}. Not paying particular attention to such a treasure trove of existing knowledge seems not only inefficient, but also potentially unsafe. The cost of a mistake in movie and restaurant recommendations is not very high; maybe one loses a couple of hours and a few hundred dollars. However, guessing the wrong decision about a technical operation in a chemical or nuclear plant could lead to potentially dangerous results. \\

These concerns underscore the need for LLMs to evolve beyond their current capabilities and incorporate both "algebraic" (i.e., symbolic) and "geometric" representations of the world, particularly for science and engineering. Such hybrid AI systems will be more reliable and interpretable, and would also require fewer data to train. We call these hybrid AI systems \textit{Large Knowledge Models} (LKMs) because they will not be limited to NLP-based techniques or NLP-like applications only. \\

The main challenges that plague current LLMs have been around in AI since the era of expert systems. As heuristic rule-based expert systems emerged in the early 1980s - they were the "LLMs" of that era - critiques observed their main failings such as their shallow understanding, limited generalization, brittleness of behavior, lack of reasoning transparency, poor inference, unreliability, etc.~\cite{venkat1988two-tier}. We find the same old issues resurfacing and some new ones due to the probabilistic nature of the new models~\cite{lenat2023LLM-CYC} in the context of LLMs. \\

Since the dawn of AI in the 1950s, there have been two opposing camps, symbolic vs. connectionist, battling for their ideas. For nearly five decades, symbolic AI dominated the discourse, culminating in the success of expert systems in the 1980s-90s. The connectionist paradigm was given a rebirth with the rediscovery of the backpropagation algorithm in 1986, and with the advent of deep neural networks around 2010~\cite{lecun2015deeplearning} it has taken off. Now, the pendulum seems to have swung to the other extreme. AI is now dominated by the connectionist paradigm. \\

Although the connectionist camp seems to have won the latest battle for ideas, we believe that, in the long run, AI will be determined by the successful integration of both the symbolic and connectionist ideas. We need the symbolic paradigm to capture first-principles-based mechanistic "deep" knowledge and reasoning, and connectionist representations for data-driven empirical knowledge~\cite{venkat1988two-tier, mann2021predicting,mann2022hybrid, mann2023susie,lenat2023LLM-CYC,venkat2024quovadis}. \\

The importance of using such human expertise has become evident even in non-technical areas for LLMs. For example, ChatGPT uses human experts' guidance as reinforcement learning using human feedback (RLHF). This is reminiscent of the techniques employed in expert systems. Another example of this realization is the development of \emph{AlphaGeometry}, which proves mathematical theorems at the Olympiad level~\cite{trinh2024solving}. As a neuro-symbolic system, it uses a language model in conjunction with a symbolic deduction engine.  

\section{More is Different}

In 1972, the physics Nobel laureate Philip Anderson published an influential paper entitled "More is Different"\cite{anderson1972more}. In it, he observes: 
\begin{quote}
"The constructionist hypothesis breaks down when confronted with the twin difficulties of scale and complexity. The behavior of large and complex aggregates of elementary particles, it turns out, is not to be understood in terms of a simple extrapolation of the properties of a few particles. Instead, at each level of complexity entirely new properties appear, and the understanding of the new behaviors requires research which I think is as fundamental in its nature as any other. That is, it seems to me that one may array the sciences roughly linearly in a hierarchy, according to the idea: The elementary entities of science X obey the laws of Y. .....

But this hierarchy does not imply that science X is "just applied Y." At each stage entirely new laws, concepts, and generalizations are necessary, requiring
inspiration and creativity to just as great a degree as in the previous one. Psychology is not applied biology, nor is biology applied chemistry."
\end{quote}

In this sense, Newtonian mechanics and $F = ma$ can explain the dynamics of a few particles. However, when we have Avogadro's number ($6.02\times 10^{23}$) of molecules dynamically interacting in a gas, the collective behavior cannot be understood by applying Newton's law $10^{23}$ times! Sure, $F = ma$ is going on at the molecular level, but there is much more happening that cannot be understood by Newton's Second Law alone. Physics does not scale this way. \\

To explain the macroscopic phenomena, we need entirely new concepts, such as density, temperature, pressure, entropy, chemical potential, etc., to predict and explain the behavior of a gas. These concepts are absent at the individual particle level in Newtonian mechanics. We need an entirely new conceptual and mathematical framework, called statistical mechanics, to address this new physics. \\

Similarly, in our opinion, when Open AI transitioned from GPT-1 to GPT-3.5 (and beyond), they created an entirely different beast that is \textit{qualitatively}, not just quantitatively, different from GPT-1. Once again, invoking a physics analogy, it is similar to the transformation from a gas (e.g., the density of steam at 1 atm and $100^{\circ}$ is 0.0006 g/cc) to a liquid (e.g., the density of water at room temperature and 1 atm is 1.0 g/cc) as the density increased by approximately 1000x. One sees a similar thousand-fold increase in parameter size from GPT-1 to GPT-3.5. A liquid is not just a scaled version of a gas. It has entirely different physics. \\

Likewise, the ultra-large language models, such as GPT-3.5 and beyond, are not mere autocomplete engines, but they have new emergent capabilities that require creating a new conceptual framework similar to the transition from Newtonian mechanics to statistical mechanics. As noted, in statistical mechanics, $F = ma$ is happening in molecular collisions, but the overall physics is much more than and much different from what is implied by Newton's laws. Similarly, in GPT-3.5 and beyond, autocomplete is happening at the individual token level, but the overall behavior is much more and much different. They may not have developed a human-like understanding of their domain, but they seem to have acquired a different kind of understanding and intelligence. It may be incomplete and flawed, but we do not yet know enough about their internals (or, for that matter, \textit{our own} internals) to categorically deny their "understanding" and "intelligence". \\

In conclusion, generative AI has opened up unimagined possibilities in all aspects of human endeavor. However, to harness its potential safely and effectively, we need to go beyond large language models to large knowledge models, which deeply incorporate fundamental knowledge and human expertise. This requires integrating the "geometric" and "algebraic" representations of the world in a new mathematical framework of large knowledge models. Obviously, nature has figured out how to do this using deep neural networks in our brains. Our challenge is to do the same in generative AI systems.

\bibliography{references}
\end{document}